\documentclass[letterpaper, 10 pt, conference]{template/ieeeconf}

\IEEEoverridecommandlockouts                              

\usepackage{graphicx}
\usepackage{caption}
\usepackage{amsmath,amssymb,enumerate}

\usepackage{amsthm}
\usepackage{mathtools}
\usepackage{breqn}
\usepackage{algorithm, algorithmicx, algpseudocode}

\usepackage{blindtext}
\usepackage{gensymb}
\usepackage{xparse}
\usepackage{lipsum}
\usepackage{mathrsfs}
\usepackage[mathscr]{euscript}
\usepackage{times}
\usepackage{cite} 
\usepackage{multicol}
\usepackage[caption=false,font=footnotesize]{subfig}
\usepackage{amsfonts}
\usepackage[utf8]{inputenc}
\usepackage[T1]{fontenc}
\usepackage{textcomp}
\usepackage{amsfonts}
\usepackage{soul}
\usepackage{multirow}
\usepackage{booktabs}
\usepackage[usenames,dvipsnames,svgnames,table, xcdraw]{xcolor}

\usepackage[colorlinks=true,pdfpagemode=UseNone,citecolor=black,linkcolor=black,urlcolor=BrickRed]{hyperref}

\usepackage{comment}

\renewcommand{\thefigure}{\arabic{figure}}

\DeclareDocumentCommand{\vectorToSkew}{ O{} }{\left(#1\right)_\times}

\usepackage{rotating}
\usepackage{outlines}
\usepackage{multirow}
\usepackage{caption}

\title{\LARGE \bf
Monocular 3D Vehicle Detection Using Uncalibrated Traffic Cameras through Homography
}

\author{Minghan Zhu$^{1}$, Songan Zhang$^{1}$, Yuanxin Zhong$^{1}$, Pingping Lu$^{1}$, \\ Huei Peng$^{1}$ and John Lenneman$^{2}$
\thanks{*This work was supported by the Collaborative Safety Research Center at the Toyota Motor North America Research \& Development.}%
\thanks{$^{1}$M.~Zhu, S.~Zhang, Y.~Zhong, P.~Lu, and H.~Peng are with the University of Michigan, Ann Arbor, MI 48109, USA. {\tt\small\{minghanz, songanz, zyxin, pingpinl, hpeng\}@umich.edu}}%
\thanks{$^{2}$J.~Lenneman is with the Collaborative Safety Research Center at the Toyota Motor North America Research \& Development, Ann Arbor, MI 48105, USA. {\tt\small john.lenneman@toyota.com}}%
}

\definecolor{mypink1}{rgb}{0.858, 0.188, 0.478}
\usepackage[dvipsnames]{xcolor}
\begin{document}

\maketitle
\thispagestyle{empty}
\pagestyle{empty}

\begin{abstract}
This paper proposes a method to extract the position and pose of vehicles in the 3D world from a single traffic camera. Most previous monocular 3D vehicle detection algorithms focused on cameras on vehicles from the perspective of a driver, and assumed known intrinsic and extrinsic calibration. On the contrary, this paper focuses on the same task using uncalibrated monocular traffic cameras. We observe that the homography between the road plane and the image plane is essential to 3D vehicle detection and the data synthesis for this task, and the homography can be estimated without the camera intrinsics and extrinsics. We conduct 3D vehicle detection by estimating the rotated bounding boxes (r-boxes) in the bird's eye view (BEV) images generated from inverse perspective mapping. We propose a new regression target called \textit{tailed~r-box} and a \textit{dual-view} network architecture which boosts the detection accuracy on warped BEV images. Experiments show that the proposed method can generalize to new camera and environment setups despite not seeing imaged from them during training. 
\end{abstract}

\section{INTRODUCTION}

Traffic cameras are widely deployed today to monitor traffic conditions especially around intersections. Camera vision algorithms are developed to automate various tasks including vehicle and pedestrian detection, tracking \cite{fedorov2019traffic}, and re-identification \cite{khan2019survey}. However, most of them work in the 2D image space. In this paper, we consider the task of monocular 3D detection, which is to detect the targets and to estimate their positions and poses in the 3D world space from a single image. It could enable us to better understand the behaviors of the targets from a traffic camera.

Monocular camera 3D object detection is a non-trivial task since images lack depth information. A general strategy is to leverage the prior of the sizes of the objects of interest and the consistency between 3D and 2D detections established through extrinsic and intrinsic parameters. To help improve the performance, a series of datasets are published with 3D object annotation associated with images. \cite{nuscenes2019, sun2020scalability, pham20203d} are datasets in driving scenarios, and \cite{ahmadyan2020objectron} contains object-centric video clips in general daily scenarios. 

However, these research efforts mostly cannot be directly applied to traffic cameras for two reasons. First, the intrinsic/extrinsic calibration information of many cameras are not available to users. Second, 3D annotations of images from these traffic cameras are lacking, while there are some with 2D annotations \cite{snyder2019streets, shah2018accident, zhang2017understanding}. Some previous work tried to solve the 3D object detection problem, but they posed some strong assumptions such as known intrinsic/extrinsic calibration \cite{zhang2020vehicle} or fixed orientation of the objects \cite{kocur2020detection}. We extend the 3D detection to a more general setup without these assumptions. 

\begin{figure}
    \centering
    \includegraphics[width=\columnwidth]{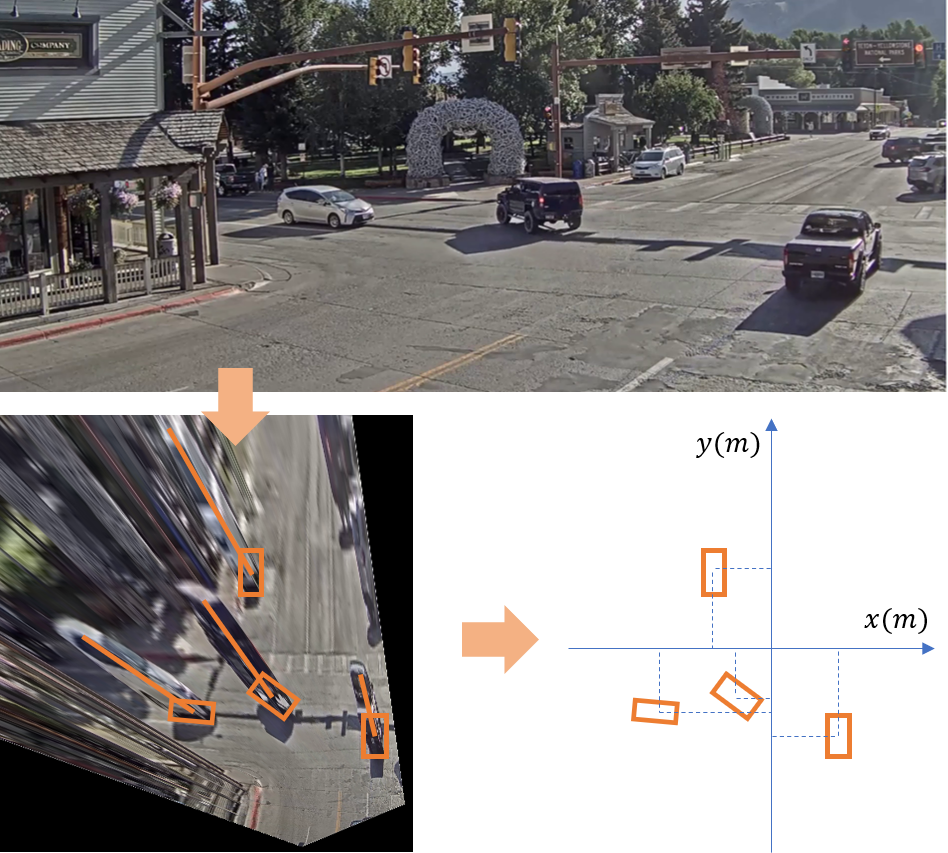}
    \caption{The 3D vehicle detection problem is transformed to a 2D detection problem in warped bird's eye view (BEV) images. The orange lines attached to each orange boxes are \textit{tails}, defined in Sec. \ref{sec:tailed r-box} and Fig. \ref{fig:tailed rbox}, which are regressed by the network to better handle distortions in BEV images. }
    \label{fig:title}
\end{figure}

We leverage the homography between the road plane and image plane as the only connection between the 3D world and the 2D images. The homography can be estimated conveniently using satellite images from public map service. As opposed to 3D bounding box detection which requires full calibration, we formulate the 3D object detection problem as the detection of rotated bounding boxes in images from bird's eye view (BEV) generated using the homography, see Fig. \ref{fig:title}. The homography also enables us to synthesize images from the perspective of a traffic camera even if it is not calibrated, which in return benefits the training of the detection network. To address the problem of shape distortion introduced by the inverse perspective mapping (IPM), we designed an innovative regression target called \textit{tailed r-box} as an extension to the conventional rotated bounding boxes, and introduced a dual-view network architecture. 

The main contributions of this paper include:
\begin{enumerate}
    \item We propose a method to estimate the pose and position of vehicles in the 3D world using images from a monocular uncalibrated traffic camera.
    \item We propose two strategies to improve the accuracy of object detection using IPM images: (a) tailed r-box regression, (b) dual-view network architecture. 
    \item We propose a data synthesis method to generate data that are visually similar to images from an uncalibrated traffic camera. 
    \item Our work is open-sourced and software is available for download at \url{https://github.com/minghanz/trafcam_3d}. 
\end{enumerate}

The remainder of this paper is organized as follows. The literature review is given in Sec. \ref{sec:review}. The proposed method for 3D detection is introduced in Sec. \ref{sec:method}. The dataset used for training and the data synthesis method are introduced in Sec. \ref{sec:data}. The experimental setup and results are presented in Sec. \ref{sec:experiments}. Section~\ref{sec:conclusions} concludes the paper and discusses future research ideas.

\section{Related Work} \label{sec:review}

\subsection{Monocular 3D vehicle detection} \label{sec:review_mono3d}
A lot of work has been done in monocular 3D vehicle detection. Our primary application is vehicle detection. Although the problem is theoretically ill-posed, most vehicles have similar shapes and sizes, allowing the network to leverage such priors jointly with the 3D-2D consistency determined by the camera intrinsics. For example, \cite{chabot2017deep} employed CAD models of vehicles as priors.  \cite{qin2019monogrnet} estimated the depth from the consistency of the 2D bounding boxes and the estimated 3D box dimensions. The object depth can also be estimated using a monocular depth network module \cite{he2019mono3d++}. Some work proposed to change to a space to deal with 3D detection better. For example, \cite{wang2019pseudo} back-projected 2D images to 3D space using estimated depth and detect 3D bounding boxes in the 3D space directly. \cite{kim2019deep} transformed original images to the bird's eye view (BEV) where vehicles can be localized with 2D coordinate, which is similar to our work, but they did not address the challenges caused by distortion in perspective transform. In this paper, we identify these challenges and propose new solutions. 

\subsection{Calibration and 3D vehicle detection for traffic cameras}
Some previous work aimed at solving the 3D detection problem from traffic cameras, but with different setup and assumptions. The detection approach is closely coupled with the underlying calibration method, as the latter determines how to establish the 3D-2D relation. Therefore we review the detection and the calibration methods together. A common type of calibration methods is based on vanishing point detection. 

Methods are proposed to detect vanishing points (VPs) from the major direction of vehicle movement and edge-shaped landmarks in the scene \cite{sochor2017traffic,you2016accurate}. Depending on whether two \cite{dubska2014fully} or three \cite{zhang2012practical} VPs are estimated, the intrinsic matrix and the rotational part of the extrinsic matrix can be solved with or without known principle point. The translation estimation requires real-world scale information, which can be obtained from, for example, known camera height \cite{zhang2012practical}, distance between two projected points \cite{dubska2014fully}, or average size of vehicles \cite{sochor2017traffic}. With the calibration, 3D bounding boxes can be constructed from 2D bounding box detection or instance segmentation following the direction of vanishing points \cite{dubska2014automatic}. There are two limitations of this approach. First, the calibration requires a lot of parallel landmarks and/or traffic flow in one or two dominant directions, which may not be the case in real traffic, e.g., roundabouts. Second, the construction of 3D bounding boxes assumes that all vehicles are largely aligned in the direction of the vanishing lines, which is not always true, including at curved lanes, intersections with turning vehicles, and roundabouts. 

\cite{zhang2020vehicle} avoided the limitations mentioned above, by calibrating the 2D landmarks in images to 3D landmarks in Lidar scans, obtaining full calibration of the camera. It is apparently non-trivial to obtain Lidar scans for already-installed traffic camera, which limits the practicality to apply this approach. The authors synthesized images of vehicles from CAD models on random background images as the training data. We adopt a similar approach, but we render the vehicles on the scene of the traffic cameras directly, which reduces the domain gap, while not requiring the intrinsic/extrinsic calibration of the cameras. 




\begin{figure*}[t]
    \centering
    \includegraphics[width=\textwidth]{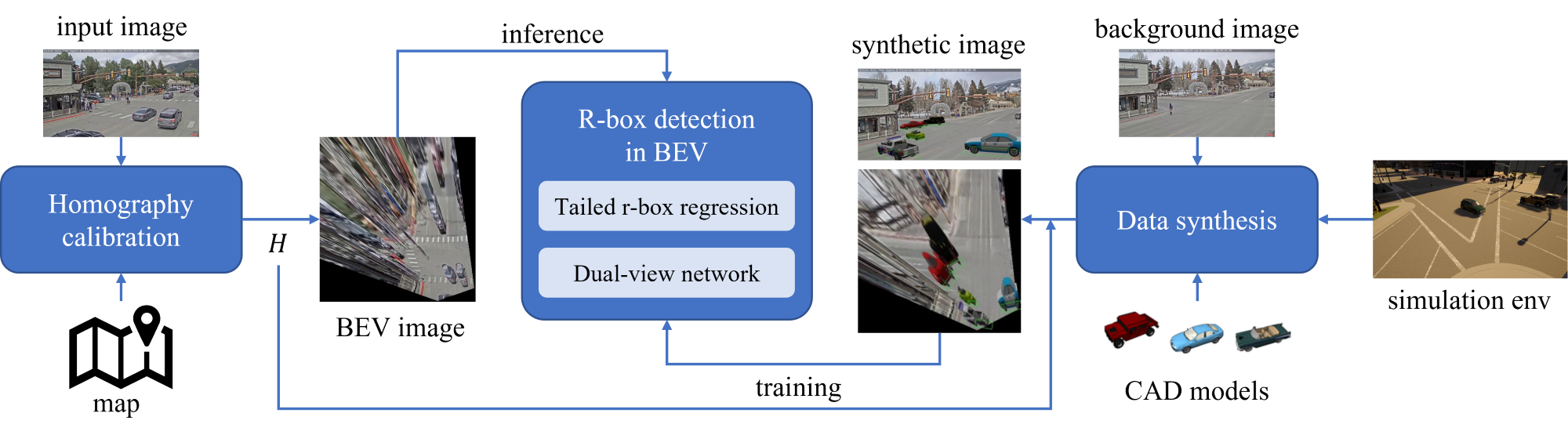}
    \caption{Overview of the 3D vehicle detection framework. }
    \label{fig:overview}
\end{figure*}

\subsection{Rotated bounding box detection}

Rotated bounding box detection is useful in aerial image processing, where objects generally are not aligned to a dominant direction as in our daily images, and are mostly axis-aligned to the gravity direction. Some representative work include \cite{zhong2020single, yang2019scrdet}. Although the regression target is very similar in this paper, the challenges are very different in that we deal with BEV images warped from the perspective of traffic cameras through IPM, which introduces severe distortion compared with original traffic camera images, and have more occlusions compared with native bird's eye view images (e.g., aerial images). 

\subsection{Perception networks with perspective transform}
Several previous work also employed a similar idea of using perspective transform to conduct perception for BEV images. \cite{kim2019deep} conducted object detection in warped BEV images, but it did not address the distortion effect in IPM, as mentioned at the end of Sec. \ref{sec:review_mono3d}. \cite{reiher2020sim2real, loukkal2021driving} addressed the distortion effect, but the discussions there are in the context of segmentation. \cite{garnett20193d, reiher2020sim2real} studied lane detection and segmentation, respectively.  They employed perspective transform inside the network, which approach we adopt. The difference is that they mainly used it to transform results to BEV, while in this work the warping is to fuse features from the original view images and the BEV images. 

\section{Proposed Method} \label{sec:method}
Our strategy is to transform the 3D vehicle detection problem to the 2D rotated bounding box detection problem in the bird's eye view. We are mainly concerned about the planar position of vehicles, and the vertical coordinate in the height direction is of little importance since we assume a vehicle is always on the (flat) ground. Under the moderate assumption of flat-Earth, and that the nonlinear distortion effect in the camera is negligible, the pixel coordinates in the bird's eye view images are simply a scaling of the real world planar coordinates. 

There are several other merits to work on bird's eye view images. First, the rotated bounding boxes of the objects at different distances have a consistent scale in the bird's eye view images, making it easier to detect remote objects. Second, the rotated bounding boxes in the bird's eye view do not overlap with one another, as oppose to 2D bounding boxes in the original view. Nevertheless, working on bird's eye view images also requires us to address the challenges of distortion and occlusion, as mentioned above, and they will be discussed in more details below in this section. 

An overview of the proposed method is shown in Fig. \ref{fig:overview}. It has three parts: homography calibration, vehicle detection in warped BEV images, and data synthesis. The data synthesis part is for network training and not directly related to the detection methodology, therefore introduced later in Sec. \ref{sec:data}. In this section, the first two parts are introduced. 

\subsection{Calibration of homography}\label{sec:calib}
A planar homography is a mapping between two planes which preserves collinearity, represented by a 3*3 matrix. We model the homography between the original image and the bird's eye view image as a composition of two homographies: 
\begin{equation}
    H_{ori}^{bev} = H^{bev}_{world} H^{world}_{ori}
\end{equation}
where $s p_a = H^a_b p_b$, denoting that $H^a_b$ maps coordinates in frame $b$ to coordinated in frame $a$ up to a scale factor $s$, and $p = [x, y, 1]^T$ is the homogeneous coordinate of a point in a plane. $bev$ denotes the BEV image plane. $world$ denotes the road plane in the real world. $ori$ denotes the original image plane. $H^{bev}_{world}$ can be freely defined by users as long as it is a similarity transform, preserving the angles between the real-world road plane and the bird's eye view image plane. Calibration is needed for $H^{world}_{ori}$, denoting the homography between the original image plane and road plane in the real world. If the intrinsic and extrinsic parameters of a camera is known or can be calibrated using existing methods, the homography can be obtained following Eq. \ref{eq:homography to inex}. Under circumstances where the full calibration is unavailable, the homography can be estimated if the corresponding points in the two planes are known. We find the corresponding points by annotating the same set of landmarks in the traffic camera image and in the map (e.g. Google Maps). Using the satellite images on the map, we can retrieve the real world coordinate of the landmarks given a chosen local frame. With the set of corresponding points $\{(p_i^{world}, p_i^{ori})\}$, the homography $H^{world}_{ori}$ can be solved by Direct Linear Transformation (DLT). Given $H_{ori}^{bev}$, the original traffic camera images can be warped to BEV images. 

\begin{figure}[t]
    \centering
    \includegraphics[width=\columnwidth]{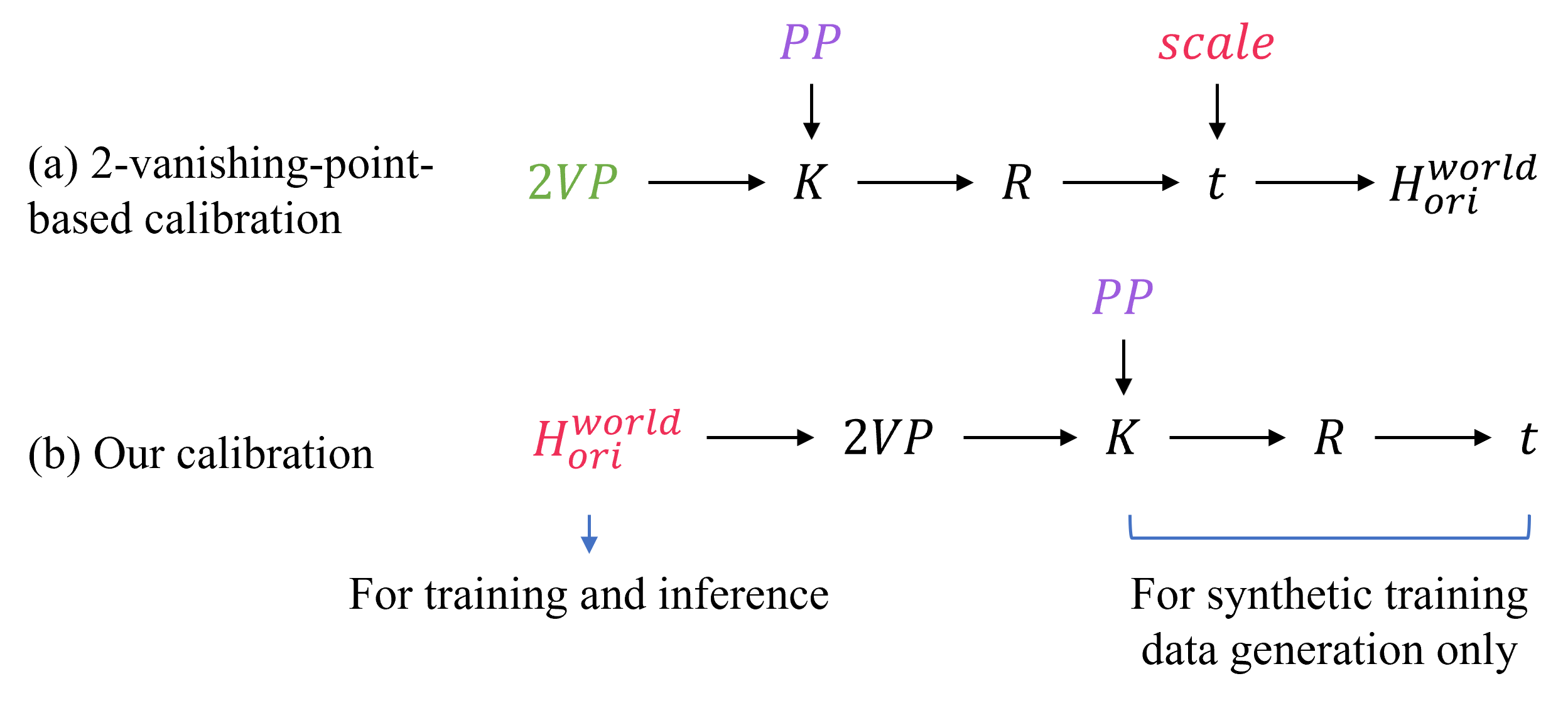}
    \caption{Comparison of our calibration (Sec. \ref{sec:calib}) and 2-vanishing-point-based calibration \cite{dubska2014fully}. \textit{PP} stands for the principal point, and \textit{VP} stands for vanishing points. \textcolor{ForestGreen}{\textit{Green}}: values that can be obtained from automatic algorithms. \textcolor{mypink1}{\textit{Pink}}: values obtained from manual point annotation. \textcolor{BlueViolet}{\textit{Purple}}: values assumed to be known in prior or specified by users (see Sec. \ref{sec:synthetic_calib} for more details). \textit{Black}: values derived from other values. }
    \label{fig:calib_comp}
\end{figure}

There are alternative ways to calibrate the homography. For example, one may also obtain $H^{world}_{ori}$ from vanishing point (VP) estimation, following \cite{dubska2014fully} for example, which can be automated. However, in order to recover the scale, real-world distance between points on image is needed. Unless camera height is known in prior, scale recovery involves annotating $\{(p_i^{world}, p_i^{ori})\}$ pairs on the map and on the traffic camera image as in our method. Therefore, we directly choose point annotation over VP estimation to calibrate $H^{world}_{ori}$. Given $H^{world}_{ori}$, the 2 VPs and scaling are also determined. See Fig. \ref{fig:calib_comp} for a comparison between VP-based calibration and ours. Notice that in our network training and inference, only $H^{world}_{ori}$ is needed. The full intrinsic ($K$) and extrinsic ($R, t$) parameters are only needed when generating synthetic training data, for which we also need to know the principal point. More detail on the latter part is in Sec. \ref{sec:synthetic_calib}. 

\subsection{Rotated bounding box detection in warped bird's eye view (BEV) images}
The rotated bounding box detection network in this paper is developed based on YOLOv3 \cite{redmon2018yolov3}, by extending it to support rotation prediction. We will abbreviate "\textit{rotated bounding box}" as "\textit{r-box}" in the following. We choose YOLOv3, which is a one-stage detector, over two-stage detectors (e.g. \cite{cai2018cascade}) for the following two reasons. First, two-stage detectors have advantage in detecting small objects and overlapping objects in crowed scenes, while in bird's eye view images the size of objects does not vary too much, and the r-boxes are not overlapping. Second, one-stage detectors are faster. More recent network architectures like \cite{bochkovskiy2020yolov4} should also work. 

The network is extended to predict rotations by introducing the yaw ($r$) dimension in both anchors and predictions. The anchors are now of the form $(l, w, r)$, where $r\in [0, \pi]$, implying that we are not distinguishing the front end and rear end of vehicles in the network. Although the dimension of the anchors increased by one, we do not increase the total number of anchors, due to the fact that object size does not vary too much in our bird's eye view images. There are 9 anchors per YOLO prediction layers, and there are in total 3 YOLO layers in the network, the same as in YOLOv3. The rotation angles of 9 anchors in a YOLO prediction layer are evenly distributed over the $[0, \pi]$ interval. 

\begin{figure}[t]
    \centering
    \includegraphics[width=\columnwidth]{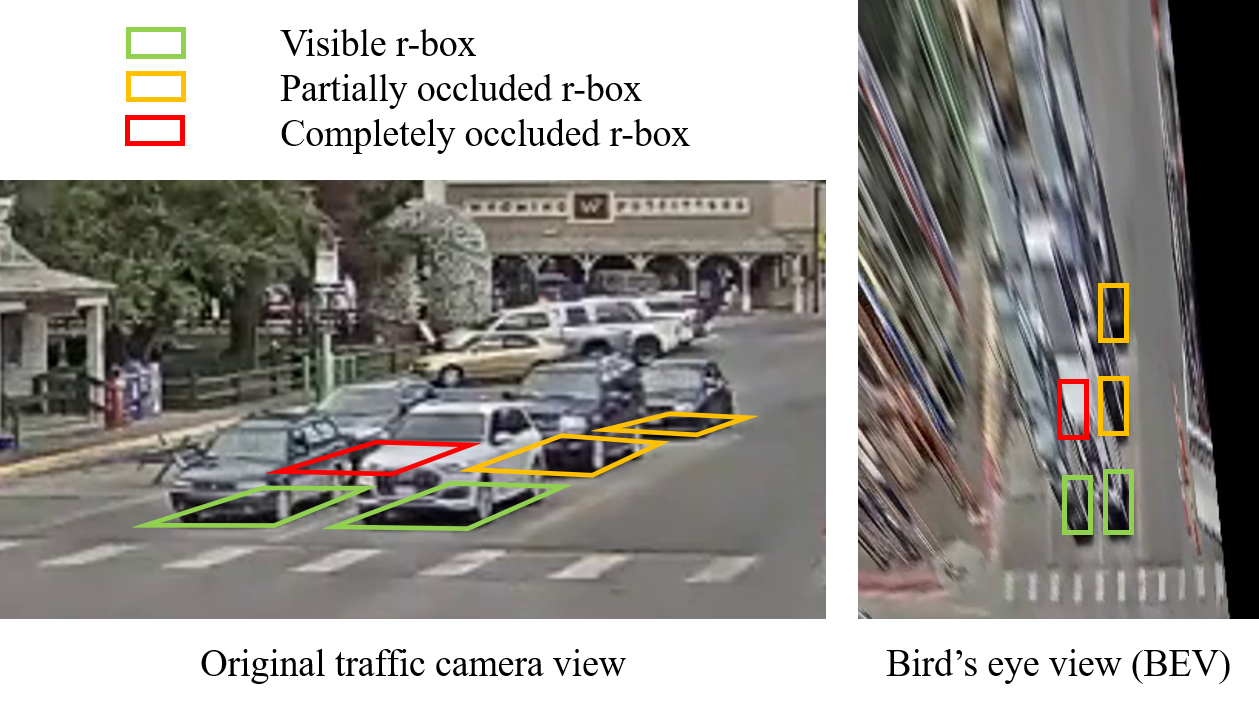}
    \caption{R-boxes shown in BEV and original camera view. The red r-box is completely occluded by surrounding vehicles, posing challenges for the network detection. Notice that there are still visible pixels of the vehicle corresponding to the red r-box, therefore detecting it is possible, but needs specific solutions (see Sec. \ref{sec:tailed r-box}). }
    \label{fig:rbox_occ}
\end{figure}

The network predicts the rotational angle offsets to the anchors. Denote the angle of an anchor as $r_0$, only anchors with $|r_0~-~r_{gt}|<\pi/4$ can be considered as positive, and for a positive anchor the rotation angle is predicted following Eq. \ref{eq:r_pred}. 
\begin{equation} \label{eq:r_pred}
    r_{pred} = \frac{\pi}{2}(\sigma(x)-0.5) + r_0
\end{equation}
where $x$ is the output of a convolution layer, and $\sigma(\cdot)$ is the sigmoid function. It follows that $|r_{pred} - r_0| < \pi/4$. 

The loss function for angle prediction is in Eq. \ref{r_loss}. Note that the angular residual $r_{res}=r_{pred} - r_{gt}\in (-\pi/2, \pi/2)$ falls in a converging basin of the $sin^2(\cdot)$ function. 
\begin{equation} \label{r_loss}
    L_{rotation} = sin^2(r_{res}) = sin^2(r_{pred} - r_{gt})
\end{equation}

\subsection{Special designs for detection in warped BEV images}

With the above setup, the network is able to fulfill the proposed task, but the distortion introduced in the inverse perspective mapping poses some challenges to the network, which harm the performance. First, in bird's eye view images, a large portion of the pixels of vehicles are outside of the r-boxes. What makes it worse, when the vehicles are crowded, the r-box area could be completely occluded and the visible pixels of the vehicle are disjoint from the r-box (see Fig. \ref{fig:rbox_occ}), which makes it difficult for the network to infer. Secondly, the IPM "\textit{stretches}" the remote pixels, extending the remote vehicles to a long shape. It requires the network to have large receptive field for each pixel to handle very large objects. Our proposed designs solve these two problems. 

\begin{figure}[t]
    \centering
    \includegraphics[width=\columnwidth]{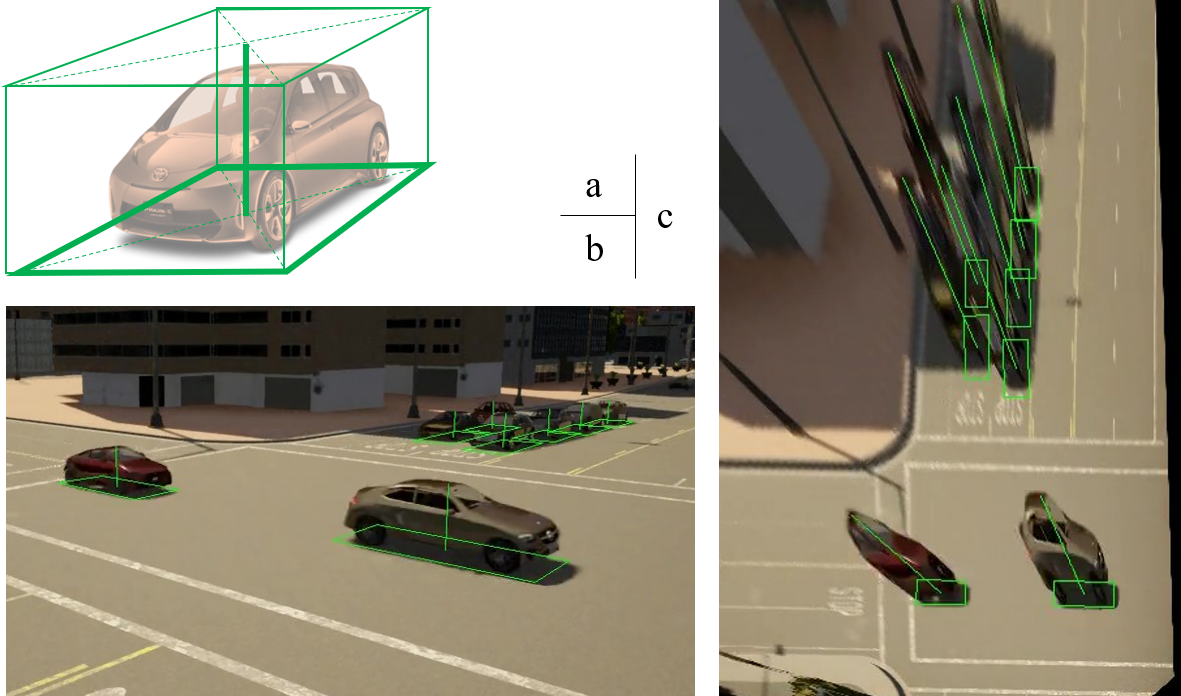}
    \caption{Definition of a \textit{tailed r-box} and examples in synthetic training data. (a) Definition of tailed r-box. The \textit{tail} is defined as the line connecting the center of the bottom face and that of the top face of the 3D bounding box. (b) Tailed r-boxes in original view. (c) Tailed r-boxes in BEV. A tail indicates the stretched pixels of a vehicle in BEV. }
    \label{fig:tailed rbox}
\end{figure}
\subsubsection{Tailed r-box regression} \label{sec:tailed r-box}
We propose a new regression target called \textit{tailed r-box} to address the problem that r-boxes could be disjoint from the visible pixels of objects. It is constructed from the 3D bounding boxes in the original view. The \textit{tail} is defined as the line connecting the center of the bottom rectangle to that of the top rectangle of the 3D bounding box. After warping to BEV, the tail extends from the r-box center through the \textit{stretched} body of the vehicle, as shown in Fig. \ref{fig:tailed rbox}. Note that while the definition of tails is in the original view images, the learning and inference of tails can be done in the BEV images. In BEV images, predicting tailed r-boxes corresponds to augmenting the prediction vector with two elements: $u_{tail}, v_{tail}$, representing the offset from the center of r-box to the end of tail in BEV. Anchors are not parameterized with tails. 

By enforcing the network to predict the tail offset, the network is guided to learn that the stretched pixels far from the r-box are also part of the objects. Especially when the bottom part of a vehicle is occluded, the network could still detect it from the visible pixels at the top, drastically improving the recall rate (see Sec. \ref{sec:experiments}). In comparison, directly regressing the projection of the 3D bounding boxes in BEV can achieve similar effect in guiding the network to leverage all pixels of a vehicle, but the projected location of the four top points is harder to determine in BEV, and creates unnecessary burden for the network. 

\begin{figure}[t]
    \centering
    \includegraphics[width=\columnwidth]{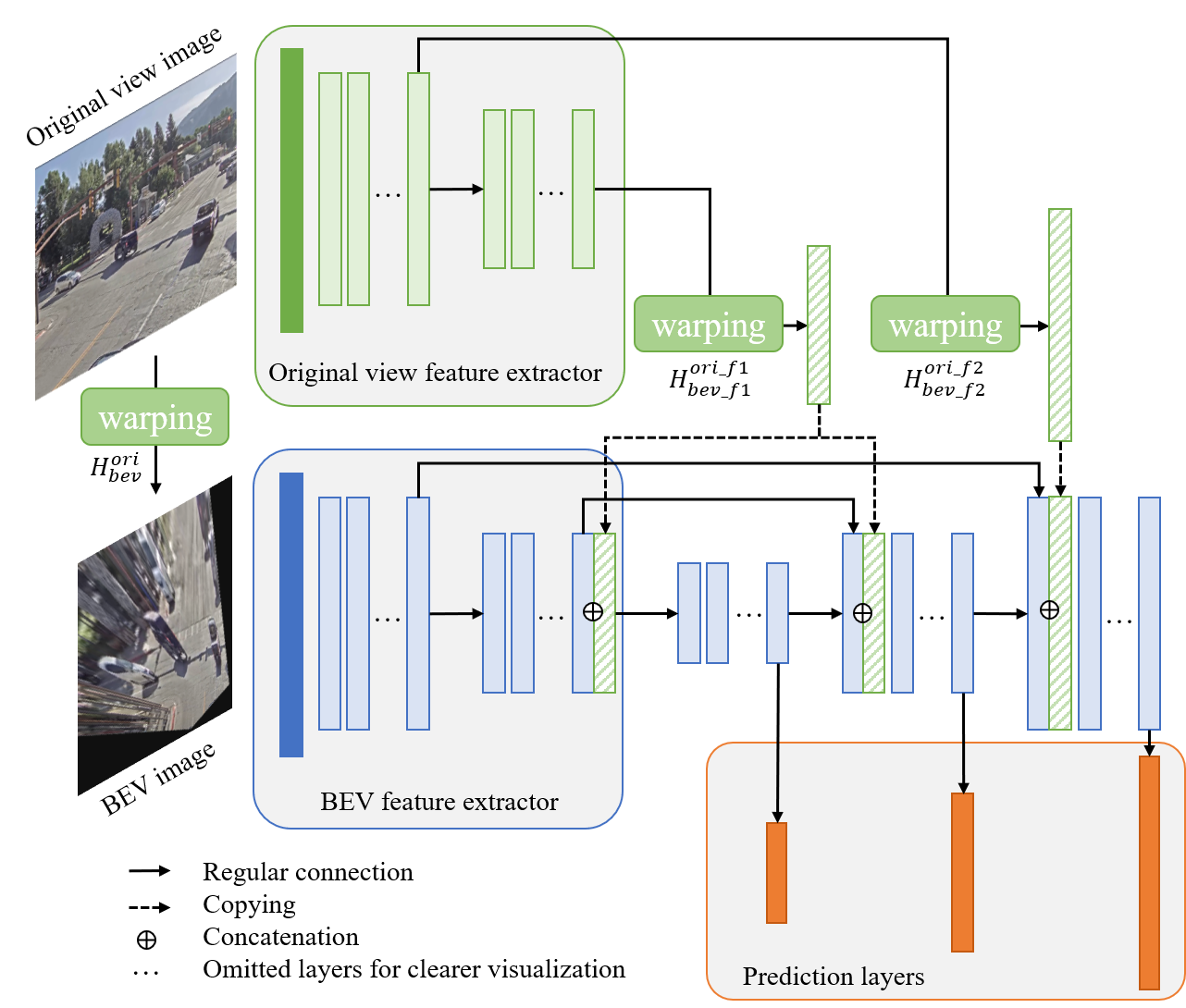}
    \caption{Dual-view network architecture. Both the original view and BEV images are taken as input. The original view feature maps are warped to BEV and concatenated with BEV feature maps. The warping (IPM) stretches the vehicles to be very long in the BEV images, posing difficulty to detection due to limited receptive field. The dual-view structure enables feature learning before warping, where the object shapes are regular and the knowledge propagation is easier.}
    \label{fig:dual view}
\end{figure}

\subsubsection{Dual-view network architecture}
The distortion in IPM makes remote objects larger than they really are, posing difficulty for learning. To alleviate the problem caused by large receptive field requirements, we propose to use a dual-view network structure. 

In the dual-view network, there are two feature extractors with identical structures and non-shared parameters, taking BEV images and corresponding original view images as input respectively. The feature maps of original images are then transformed to BEV through IPM and concatenated with the feature maps of the BEV images. The IPM of feature maps is similar to the IPM of raw images, with different homography matrices. The homography between the feature maps of original view and BEV can be calculated using Eq.~\ref{eq:homography}. 
\begin{equation} \label{eq:homography}
    H^{bev\_f}_{ori\_f} = H^{bev\_f}_{bev}H^{bev}_{ori}H^{ori}_{ori\_f}
\end{equation}
where $H^{bev\_f}_{bev}$ and $H^{ori}_{ori\_f}$ denotes the homography between the input image coordinates and the feature map coordinates, which are mainly determined by the pooling layers and convolution layers with strides. The network structure is shown in Fig.~\ref{fig:dual view}.

With the dual-view architecture, pixels of a vehicle are spatially closer in the original view images than in the BEV images, making it easier to propagate information among the pixels. Then the intermediate feature warping \textit{stretches} the information with IPM, propagating the consensus of nearby pixels of an object in the original view to pixels of further distances in BEV. In the experiments we show that the dual-view architecture improves the detection performance. 

\section{Data synthesis} \label{sec:data}
The lack of training data for 3D vehicle detection in traffic camera images pose difficulty in learning a high-performance detector. In this work, we adopt two approaches to synthesize training data. 

\subsection{CARLA-synthetic} 
Our first approach is to generate synthetic data using a simulation platform CARLA \cite{Dosovitskiy17}. CARLA is capable of producing photo-realistic images from cameras with user-specified parameters. It is able to simulate different lighting conditions and weather conditions. It also supports camera post-processing effects, e.g., bloom and lens flares. We selected several positions in the pre-built maps and collected images from the perspective of traffic cameras. 
\subsection{Blender-synthetic}\label{sec:synthetic_calib} 
The second approach is to synthesize images by composing real traffic scene background images with rendered vehicle foreground from CAD models. The background images are pictures of empty road taken by traffic cameras. The rendering and composition is by using the 3D graphic software Blender. While the CARLA images presents large varieties which benefits generalization, the discrepancy between synthesized images and real images is still easily perceivable from human eyes. Composing real background images with synthesized vehicle foregrounds could be a step forward minimizing the domain gap. A key question is: how to set up the camera in foreground rendering, such that when compositing the foreground and the background images together, the output looks like the foreground vehicles are laying on the ground, instead of floating in the air, despite that the intrinsic and extrinsic parameters corresponding to the background images are unknown? 

\begin{figure}[t]
    \centering
    \includegraphics[width=\columnwidth]{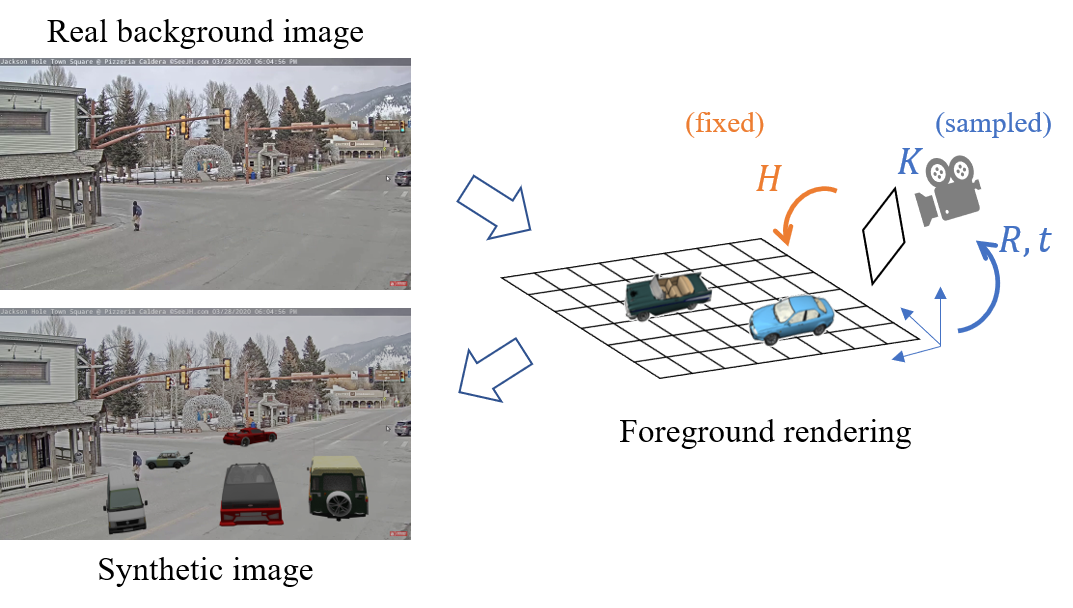}
    \caption{Synthesizing images with real background captured by traffic cameras and rendered vehicles using CAD models. The intrinsic/extrinsic parameters corresponding to the background images are unknown, but we can still render visually realistic images by sampling camera parameters that keep the homography $H$ invariant. }
    \label{fig:synthetic H}
\end{figure}

Our observation is that the plausibility mainly depends on the homography. In other words, if we can maintain the same homography from the road plane to the image plane in both foreground and background images, the composite images will look like the vehicles are on the ground, as seen in Fig. \ref{fig:synthetic H}. The relation between homography and camera intrinsic/extrinsics is shown in Eq. \ref{eq:homography to inex}. 

\begin{equation} \label{eq:homography to inex}
    s K[r_1~r_2~t] = H
\end{equation}
where $K$ is the intrinsic matrix, $T = [r_1~r_2~r_3~t]$ is the extrinsic matrix of the camera, $r_i$ is the $i$th column of the $T$ matrix, $s$ is a scaling factor, and $H=H^{ori}_{world}$ is the homography between the original image plane and the road plane in real world. Given $H$ from calibration, $K$ and $T$ are not unique, but any pair of them qualifying the constraint of Eq. \ref{eq:homography to inex} should give visually plausible image. 

Assuming $K = [f~0~p_x;~0~f~p_y;~0~0~1]$ (zero-skew and square pixels), $K$ and $T$ can be determined by specifying the principal point $P=[p_x, p_y]$. First, a pair of vanishing points $U=[u_x, u_y]$, $V=[v_x, v_y]$ can be calculated from $H$: 
\begin{equation}
    U = [\frac{h_{11}}{h_{31}}, \frac{h_{21}}{h_{31}}], V = [\frac{h_{12}}{h_{32}}, \frac{h_{22}}{h_{32}}]
\end{equation}
where $h_{ij}$ is the element of $H$ at the $i$th row and the $j$th column. Then calculate $f$ with: 
\begin{equation}
    f=\sqrt{ -\langle U-P, V-P \rangle}
\end{equation}
With $K$ calculated, $T$ can be obtained with Eq. \ref{eq:homography to inex}. Although it is common to assume $P$ to be at the image center, it is not always true in practice. We sample the principal points around the image center to create more data variability, and use the derived $(K, T)$ pairs to generate synthetic data. 

\section{Experiments} \label{sec:experiments}

The overall setup of the experiment is training on synthetic data generated following Sec. \ref{sec:data}, and testing on real data. 

\begin{figure}[t]
    \centering
    \includegraphics[width=\columnwidth]{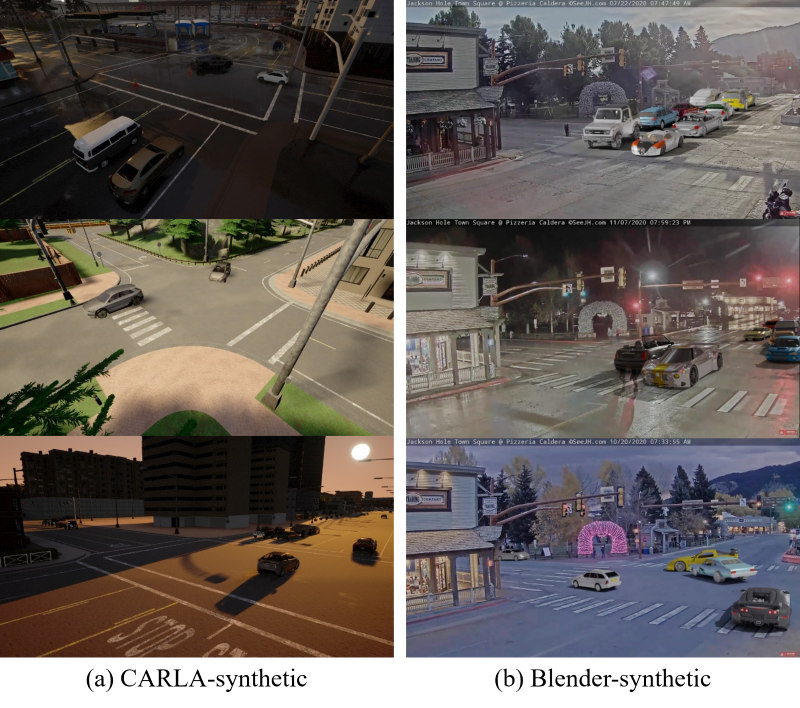}
    \caption{Examples of synthetic training data. (a) CARLA-synthetic images are captured under various weather and lighting conditions at several intersections. (b) CAD models of vehicles are rendered on real background images. Lighting effect, surface reflection, and shadow are also captured in the rendering to make the images more realistic.  }
    \label{fig:training_example}
\end{figure}

The training dataset contains 40k synthetic images consists
of two parts: CARLA-synthetic and Blender-synthetic. See Fig. \ref{fig:training_example} for some examples. The CARLA-synthetic set contains 15k images collected from 5 locations in 2 maps pre-built by CARLA, covering 2 four-way intersections, 1 five-way intersection, 1 three-way intersection, and 1 roundabout. The weather and lighting conditions are dynamically changed during the data collection, improving the robustness for adversarial environments. The Blender-synthetic set has 25k images rendered with background images from 3 traffic cameras at 2 intersections, and the CAD models for vehicle rendering are from ShapeNet \cite{shapenet2015}. 

The test dataset for quantitative evaluation contains two datasets of real videos from traffic cameras: Ko-PER \cite{koper} and BrnoCompSpeed \cite{sochor2018comprehensive}. More details of these two datasets are in Sec.~\ref{sec:quantitative}. For qualitative evaluation, we show examples of detection on the test dataset and also on traffic cameras without ground truth annotations. 


Our network is developed based on the SPP (Spatial Pyramid Pooling) variant of the YOLOv3 network. The network is trained for 20 epochs with batch size 6 on a single NVIDIA 2080Ti GPU, using ADAM optimizer and cosine learning rate decay, starting from learning rate 0.01. 

\begin{figure*}[t]
    \centering
    \includegraphics[width=\textwidth]{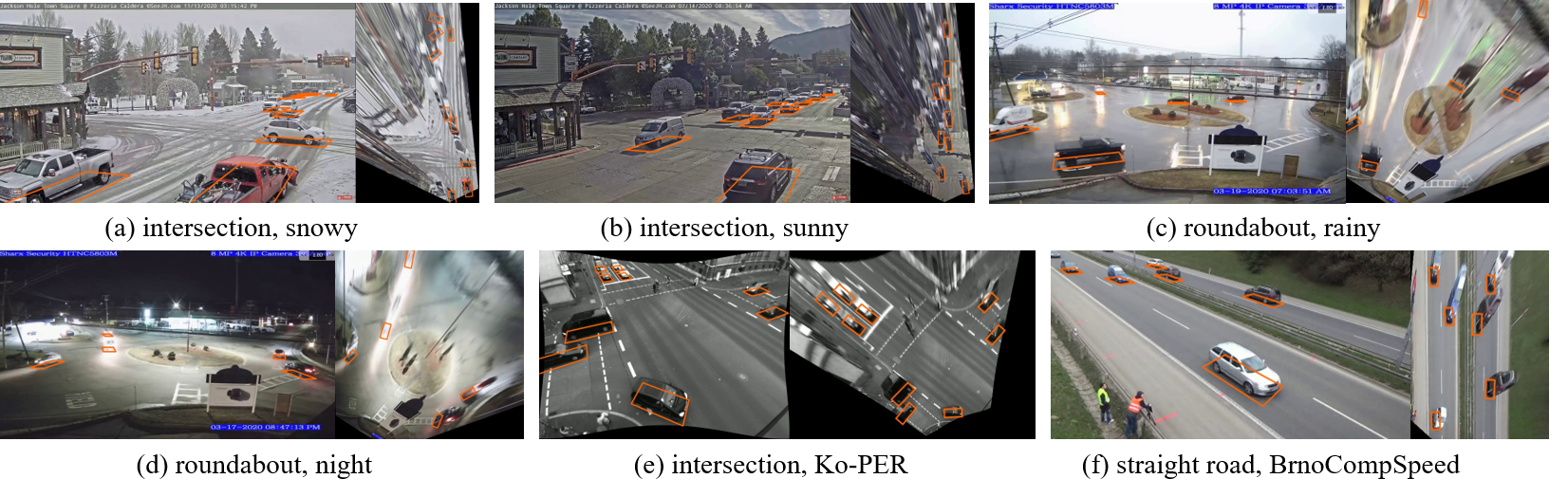}
    \caption{Examples of detection on real images in variant environment conditions. Each example contains a pair of original-view and BEV images. The vertices of detected r-boxes in BEVs are projected on the original images. (e) and (f) are samples of the quantitative test set, while (a) to (d) are from traffic cameras without ground truth annotation. Tail predictions are not drawn for better visualization. }
    \label{fig:detection_example}
\end{figure*}

\subsection{Quantitative result} \label{sec:quantitative}
\subsubsection{BrnoCompSpeed dataset}
We firstly show the quantitative evaluation on the BrnoCompSpeed dataset. The BrnoCompSpeed dataset contains eighteen 1-hour traffic videos of 50 FPS taken at 6 locations with 3 cameras per location. The videos do not have 3D bounding box annotations. Instead, the speeds and timestamps when a vehicle passes several predefined lines are recorded. Therefore, the dataset cannot be used for training a detection network, but can be used for evaluation using the protocol provided by the dataset. To be consistent with the literature, we evaluate on split C of the dataset, which contains 9 hours of videos. The result is shown in Table \ref{tab:quan_brno}. The previous work in the table are variants of methods based on vanishing point estimation and constructing 3D bounding boxes from 2D bounding box detection or segmentation. 

The key property that we examine through the evaluation on the BrnoCompSpeed dataset is the generalizability to new camera intrinsic/extrinsic parameters and environment layout, because the test set involves 9 cameras at three locations, none of which has been used for training. Therefore the environments and camera setups are completely new to the network when tested. The result shows that our network generalizes well to these new scenes, and the proposed r-box detection in BEV drastically improves the accuracy of 3D vehicle detection in terms of both recall and precision over previous methods. 

\subsubsection{Ko-PER dataset}
The Ko-PER dataset has a video of about 400 seconds corresponding to 9666 images annotated with 3D bounding boxes, captured at one intersection from two cameras with different poses. The major challenges in this dataset are the black-white images and heavy occlusions of queued vehicles at the intersection. We did not find reported results on this dataset in previous literature. Actually, the 3D detection methods based on vanishing points do not work for turning vehicles at intersections, which occur in Ko-PER dataset. 

Therefore we use this dataset for ablation study. The result is in Table \ref{tab:ablation}. We report two sets of result, evaluated using the $IoU$ with ground truth r-boxes and using the offset of center predictions to the ground truth r-boxes. The latter criteria is used to indicate the accuracy in predicting vehicle positions in the real world. Tails are not considered in the evaluation. From the table, we see that both tailed r-box targets and dual-view architecture boost the detection accuracy significantly, compared with vanilla r-box detection on BEV (which is similar to the solution in \cite{kim2019deep}). Predicting tailed r-boxes guides the network to learn long-distance connections in distorted BEV, which allows the network to better handle occluded targets. The introduction of dual-view networks further improves the accuracy. 

\begin{table}[]
\captionsetup{justification=centering}
\caption{Quantitative evaluation on BrnoCompSpeed dataset split C. The numbers of previous work are cited from \cite{kocur2020detection}. Mean precision and mean recall are calculated using the dataset evaluation protocol.}
\label{tab:quan_brno}
\begin{center}
\begin{tabular}{l|cc}
\hline
Method      & Mean recall (\%) & Mean precision (\%) \\
\hline
DubskaAuto \cite{dubska2014automatic}  & 90.08            & 73.48               \\
SochorAuto \cite{sochor2017traffic} & 83.34            & 90.72               \\
Transform3D-1 \cite{kocur2019perspective} & 89.32            & 87.67               \\
Transform3D-2 \cite{kocur2020detection} & 86.32            & 88.32               \\
Ours        & \textbf{97.25}   & \textbf{92.70}     \\
\hline
\end{tabular}
\end{center}
\end{table}

\begin{table}[]
\captionsetup{justification=centering}
\caption{Ablation study, evaluated on the Ko-PER dataset. $IoU$ is defined for r-boxes. $d$ is the distance between centers of predicted and ground truth r-boxes. $l$ is the length of ground-truth rbox. $d\leq 0.5 l$ only evaluates the position prediction. }
\begin{center}
\begin{tabular}{l|cc}
\hline
Network settings         & \multicolumn{2}{c}{Average precision (AP, \%)}  \\ \cline{2-3}
 & $IoU \geq 0.5$ & $d \leq 0.5 l$ \\ \hline
r-box (similar to \cite{kim2019deep}) & 65.67 & 71.96 \\
dual-view                & 75.78 & 83.25             \\
tailed r-box             & 78.27 & 85.55             \\
tailed r-box + dual-view (ours) & \textbf{82.44} & \textbf{91.20}             \\ \hline
\end{tabular}\label{tab:ablation}
\end{center}
\end{table}

\subsection{Qualitative results}
Examples of detection on real images are shown in Fig. \ref{fig:detection_example}. All detections are by the same network, without separate retraining for different cameras. It is shown that the network generalizes to different road layout and camera poses, and it is robust to different lighting and weather conditions. Notice that the roundabout scene and the BrnoCompSpeed dataset are not used as background in synthesizing the training data, therefore is totally unseen during learning, yet the network still achieve accurate detection. 

\section{Conclusion} \label{sec:conclusions}
We developed a method to solve the problem of 3D vehicle detection from images captured by traffic cameras, without requiring the intrinsic and extrinsic parameters of the cameras. The homography to connect images to the real world is conveniently done using public map services and only requires mild flat-Earth assumption. Based on the homography, the 3D vehicle detection problem transforms to rotated bounding box detection using BEV images. We further proposed a new regression target called \textit{tailed r-box} and a \textit{dual-view} network architecture to address the distortion and occlusion problems which are common in warped BEV images. Experiments show that both the tailed r-box regression and the dual-view structure improved the accuracy significantly. We also synthesized a large dataset via two approaches, the network trained on which generalizes well on the real test data.  The large data size also may have played a role in our significantly higher detection accuracy.

Our work provides a practical and generalizable solution to deploy 3D vehicle detection on already widely available traffic cameras. Many with unknown intrinsic/extrinsic calibration. Some interesting future direction includes removing the assumptions of planar road surface and negligible camera nonlinear distortion, by incorporating a more advanced calibration procedure. Employing self-training techniques to use unlabeled real-world data may also reduce the domain gap and achieve better precision.  

\section*{ACKNOWLEDGMENT}
This article solely reflects the opinions and conclusions of its authors and not CSRC or any other Toyota entity.

\bibliographystyle{bib/IEEEtran}
\bibliography{bib/strings-abrv,bib/ieee-abrv,bib/refs_rbox}

\end{document}